\documentclass{article}

\PassOptionsToPackage{numbers,square}{natbib}

    \usepackage[preprint]{neurips_2025}

\usepackage[utf8]{inputenc} 
\usepackage[T1]{fontenc}    
\usepackage{hyperref}       
\usepackage{url}            
\usepackage{booktabs}       
\usepackage{amsfonts}       
\usepackage{nicefrac}       
\usepackage{microtype}      
\usepackage{xcolor}         

\usepackage{multirow}
\usepackage{makecell}
\usepackage{subcaption}
\usepackage{graphicx}
\usepackage{amsmath}
\usepackage{wrapfig}

\title{SignAligner: Harmonizing Complementary Pose Modalities for Coherent Sign Language Generation}

\author{%
  Xu Wang$^{1}$, Shengeng Tang$^{1}$\thanks{Corresponding author.}, Lechao Cheng$^{1}$, Feng Li$^{1}$, Shuo Wang$^{2}$, Richang Hong$^{1}$ \\
  $^{1}$Hefei University of Technology, $^{2}$University of Science and Technology of China\\
  \texttt{wangxu2002@mail.hfut.edu.cn, tangsg@hfut.edu.cn}
}

\begin{document}

\maketitle

\begin{abstract}
Sign language generation aims to produce diverse sign representations based on spoken language. However, achieving realistic and naturalistic generation remains a significant challenge due to the complexity of sign language, which encompasses intricate hand gestures, facial expressions, and body movements. In this work, we introduce PHOENIX14T+, an extended version of the widely-used RWTH-PHOENIX-Weather 2014T dataset, featuring three new sign representations: Pose, Hamer and Smplerx. We also propose a novel method, SignAligner, for realistic sign language generation, consisting of three stages: text-driven pose modalities co-generation, online collaborative correction of multimodality, and realistic sign video synthesis. First, by incorporating text semantics, we design a joint sign language generator to simultaneously produce posture coordinates, gesture actions, and body movements. The text encoder, based on a Transformer architecture, extracts semantic features, while a cross-modal attention mechanism integrates these features to generate diverse sign language representations, ensuring accurate mapping and controlling the diversity of modal features. Next, online collaborative correction is introduced to refine the generated pose modalities using a dynamic loss weighting strategy and cross-modal attention, facilitating the complementarity of information across modalities, eliminating spatiotemporal conflicts, and ensuring semantic coherence and action consistency. Finally, the corrected pose modalities are fed into a pre-trained video generation network to produce high-fidelity sign language videos. Extensive experiments demonstrate that SignAligner significantly improves both the accuracy and expressiveness of the generated sign videos.
\end{abstract}

\section{Introduction}
Sign language is both a rich visual language and a primary form of communication within the deaf community. As a result, Sign Language Generation (SLG) is gaining significant attention in the field of visual languages and has become a classical yet challenging task. SLG encompasses various representational forms, including pose, avatar, and realistic video, each emphasizing different levels of action details and semantic representations.

Early SLG works primarily focused on avatar-based methods~\cite{Baldassarri2002SSL, glauert2006vanessa}, which require expensive pose pre-acquisition costs due to rule-based phrase lookups in pre-captured databases. Given the critical role of pose in conveying the semantics of sign language, there has been a growing shift towards the study of text-to-pose generation~\cite{zelinka2019nn, cui2019deep, stoll2020text2sign, krishna2021gan, hwang2021non, xiao2020skeleton}. Inspired by the success of transformer models~\cite{unanue2021berttune, radford2021clip}, Saunders \emph{et al.} introduced a progressive transformer for end-to-end sign pose generation~\cite{saunders2020ptslp}. In a similar vein, Huang \emph{et al.}~\cite{huang2021towards} proposed a non-autoregressive model that adopts parallel decoding, addressing issues such as error accumulation and high inference latency typically encountered in autoregressive models used in previous G2P methods. LVMCN\cite{wang2024linguistics} tackles the SLG by bridging the semantic gap between modalities and addressing the lack of word-action correspondence labels, which are crucial for strong supervision and alignment. With the rapid advancements in diffusion models, Sign-IDD~\cite{tang2024signidd} constrains joint associations and gesture details through limb-skeletal modeling, significantly enhancing the accuracy and naturalness of generated poses.

With the booming development of large-scale pre-trained modeling techniques~\cite{fu2025speculative,li2022pre,hu2023survey}, research on generating sign language videos based on digital humans is becoming a cutting-edge hotspot. Saunders \emph{et al.} ~\cite{saunders2022signing} innovatively proposed the SIGNGAN framework, which is capable of mapping directly from parameterized skeletal pose sequences to high-fidelity sign language videos, realizing end-to-end integration of action generation and rendering. Xie \emph{et al.} ~\cite{xie2024sign} developed a new method for directly generating real sign language videos without the need for intermediate human pose representations by jointly training a video generator with a hidden-space feature decoder, so that the model can directly capture the dynamic sign language of temporal dependencies. However, the existing research paradigms are still limited by the shackles of modal fragmentation: they are either confined to single-modal information representation or rely on multi-stage pipeline architectures, resulting in inherent shortcomings such as reduced semantic fidelity and lack of multimodal co-evolutionary mechanisms in the sign language generation process. Specifically, the independent treatment of gesture and avatar ignores the multi-channel coupling characteristics of sign language, while the staged generation strategy will cause the attenuation of motion details due to the quantization loss of intermediate representations, which essentially severs the essential property of sign language as a spatio-temporal continuum. As shown in Table~\ref{tab:The Ground Truth on PHOENIX14T+ dataset}, real sign language videos provide better feedback on its essential semantic properties compared to any single modality \textbf{Pose}, \textbf{Hamer} and \textbf{Smplerx}, so it is crucial to establish a new sign language representation fusion model.

\begin{figure*}[t]
  \centering
  \includegraphics[width=\textwidth]{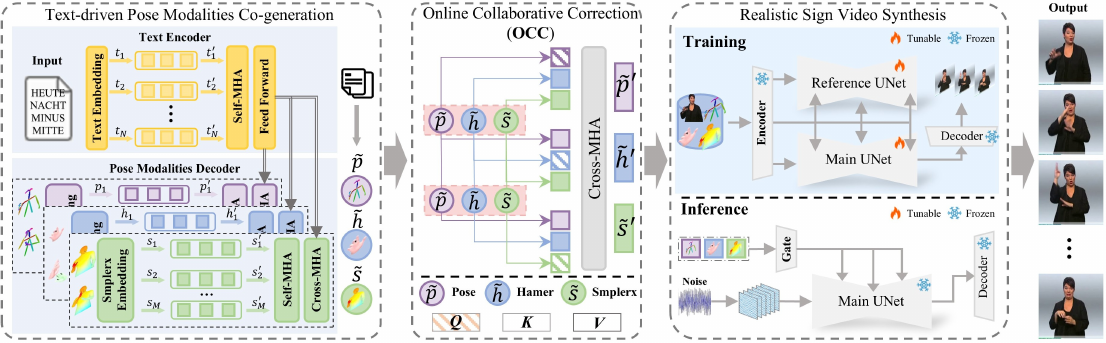}
  \caption{\small Overview of the proposed SignAligner. It contains three stages: text-driven pose modalities co-generation, online collaborative correction of multimodality, and realistic sign video synthesis. First, a joint sign language generator produces three pose modalities: $\tilde{p}_\text{1:m}$, $\tilde{h}_\text{1:m}$, $\tilde{s}_\text{1:m}$, representing posture, handshape, and body motion. Next, an online collaborative correction mechanism refines these representations, enhancing their naturalness and spatial accuracy. Finally, a photo-realistic sign language video is synthesized using a pre-trained video synthesis network.
  }
  \label{fig:main}
  \vspace{-5mm}
\end{figure*}

To this end, we propose a novel method termed \textit{SignAligner} for realistic sign language generation. As shown in Figure~\ref{fig:main}, SignAligner consists of three stages: text-driven pose modalities co-generation, online collaborative correction, and realistic sign video synthesis. Firstly, by combining text information, three sign language representation generators are used to simultaneously obtain posture coordinates, gesture actions, and body motion trajectories. In this case, the text encoder uses Transformer architecture to extract semantic features, and the generation module combines text semantic information through cross modal attention mechanism, and simultaneously generates multi-source sign language representations to ensure accurate mapping and diversity control of modal features. In online collaborative correction of multimodality, we use a cross-modal attention mechanism and a dynamic loss weighting strategy to perform online correction on the generated pose modalities, achieving information complementarity between different modalities, dynamically eliminating spatiotemporal conflicts between modalities, and ensuring semantic coherence and action consistency of the generated results. Finally, the corrected pose modalities are input into a pre-trained synthesis network to obtain high-fidelity sign videos. Our main contributions are summarized as follows:
\begin{itemize}
\item To address the limited and homogeneous nature of existing sign datasets, we have extended the widely-used sign corpus RWTH-PHOENIX-Weather 2014T~\cite{camgoz2018neural} by incorporating multiple sign representations. These include high-precision skeleton data with facial keypoints (\textbf{Pose}), hand details (\textbf{Hamer}), and 3D full-body posture (\textbf{Smplerx}), which aims to provide a novel, diverse, and high-quality resource for the community. 
\item We establish a three-stage sign video generation method, which utilizes the complementarity among pose modalities and introduces textual information to guide the generation process, thereby achieving sign language videos that can more accurately reflect the semantic content of the text.
\item We propose a joint generation mechanism that collaboratively generates sign language representations and utilizes a multimodal correction strategy combined with dynamic loss constraints to fully explore complementary information between modalities, ensuring higher quality sign language video generation in terms of semantic consistency, naturalness of actions, and spatial expression accuracy.
\end{itemize}

\begin{wraptable}{r}{0.7\textwidth}
\renewcommand\arraystretch{1.2}
\vspace{-3mm}
\caption{\small The Ground Truth of different sign language representations on PHOENIX14T+ dataset.}
\label{tab:The Ground Truth on PHOENIX14T+ dataset}
\centering
\resizebox{0.7\textwidth}{!}{
\begin{tabular}{cccccccc}
\Xhline{1pt}
   Ground Truth &BLEU-1{$\uparrow$} &BLEU-2{$\uparrow$} &BLEU-3{$\uparrow$} &BLEU-4{$\uparrow$} &ROUGE{$\uparrow$} \\
\Xhline{0.5pt}
Pose  &30.11 &20.86 &15.70 &12.50 &29.68 \\
Hamer  &31.90 &22.43 &16.68 &13.24 &31.21 \\
Smplerx &30.18 &20.40 &14.92 &11.84 &28.91 \\
\Xhline{0.5pt}
Video &38.45 &28.23 &21.65 &17.42 &37.65 \\
\Xhline{1pt}
\end{tabular}}
\vspace{-2mm}
\end{wraptable}

\section{Related work}
\subsection{Single-stage Sign Language Generation}
Sign language is both a rich visual language and a preferred mode of communication for the deaf community. With the growing need for effective communication between deaf and hearing people in recent decades, Sign Language Generation (SLG)~\cite{cui2019deep,stoll2020text2sign,saunders2020ptslp,huang2021towards,tang2024GCDM,wang2024linguistics} have received significant attention in recent years. Current work focuses on single-stage SLG, \emph{i.e.}, text-to-pose, text-to-avatar, and text-to-video.

\textbf{Text to Pose}
Sign language pose videos are widely used because they can capture the semantic and dynamic characteristics of actions through sequences of skeletal key points, have the advantage of accurately determining the orientation of sign language actions. 
Early methods were mainly based on RNN models~\cite{zelinka2019nn,cui2019deep,zelinka2020neural}, Generative Adversarial Network (GAN)~\cite{stoll2020text2sign,krishna2021gan,vasani2020generation} and Variational Auto Encoder (VAE)~\cite{hwang2021non,xiao2020skeleton}.
Inspired by the great success of transformer~\cite{unanue2021berttune,radford2021clip,tang2024GCDM,peng2025lmm,yang2023exploring}, Saunders \emph{et al.} design a progressive transformer to generate sign poses in an end-to-end manner~\cite{saunders2020ptslp}.
Huang emph{et al.} ~\cite{huang2021towards} propose a parallel decoding scheme to solve the error accumulation problem caused by autoregression in previous G2P methods. LVMCN~\cite{wang2024linguistics} solves the semantic gap between modalities and the lack of word-action correspondence labels required for strong supervised alignment in SLG. 
With the rapid development of diffusion models, Xie \emph{et al.}~\cite{xie2024g2p} ingeniously combine VAE with vector quantization to propose Pose-VQVAE, which effectively generates discrete potential representations for continuous pose sequences. Sign-IDD~\cite{tang2024signidd} enhances the accuracy and naturalness of pose generation by constraining joint associations and gesture details through skeletal modeling.

\textbf{Text to Avatar}
Early SLG experimented with avatar-based approaches due to the difficulty of meeting the demands of visual presentation with a single skeleton of information.
Baldassarri \emph{et al.}~\cite{Baldassarri2002SSL} has develop an animation engine that allows avatars to adjust their sign language and expressions according to the interpreter's mood.
Glauert \emph{et al.}~\cite{glauert2006vanessa} design the VANESSA system to allow users to convert information into virtual sign language avatars through voice or text input.
T2S-GPT~\cite{yin2024t2s} propose a dynamic vector quantization DVA-VAE model that dynamically adjusts the encoding length according to the density of the sign language information and generates a corresponding 3D avatar.
Baltatzis \emph{et al.}~\cite{baltatzis2024neural} utilizes a diffusion model combining the SMPL-X skeleton and a graph neural network to generate dynamic 3D avatar sequences from unconstrained inputs, marking a major step forward from SLG to realistic neural avatars.

\textbf{Text to Video}
Thanks to the rapid development of artificial intelligence technology, generating realistic sign language videos has gradually become possible.
Kaur \emph{et al.}~\cite{kaur2016hamnosys} develop HamNoSys, a sign language transcription system that can be directly mapped to an avatar, and each of its symbols contains a description of the initial posture and the movement over time. Xie \emph{et al.} ~\cite{xie2024sign} develop a new method to produce high-quality sign language videos without the intermediate step of human pose. It first learns from the generator and the hidden features of the video, and then uses another model to understand the order of these hidden features.

\subsection{Multi-stage Sign Language Generation}
The first deep learning-based SLG pipeline decomposes the task into three steps: first, Text-to-Gloss (T2G) translation, then second, Gloss-to-Pose (G2P) generation, and finally, Pose-to-Video (P2V) mapping~\cite{stoll2020text2sign}.
With the continuous advancement of deep learning, BROCK \emph{et al.}~\cite{brock2020learning} generate 3D kinematic skeleton data from sign language videos captured by a single camera, and then estimates the angular displacements of all the joints over time by inverse kinematics and maps them to virtual sign language characters for video generation.
Stoll \emph{et al.}~\cite{stoll2018sign} propose a realistic sign language video generation system, which firstly utilizes an encoder-decoder network to translate spoken sentences into gloss sequences, then establishes the mapping relationship between gloss and pose through a data-driven approach, and finally synthesizes a video model of sign language driven by the generated pose.
In addition, Saunders \emph{et al.} ~\cite{saunders2022signing} propose the frame selection network FS-NET as well as SIGNGAN to generate highly realistic sign language videos directly from skeleton poses.
However, these methods generally suffer from over-reliance on a single modality or limited information representation, making it difficult to fully capture the semantic and visual features of sign language. Therefore, we propose a multi-stage sign language generation method, which guides the generation of multimodal sign language representation data through text, and ultimately synthesizes sign language videos with more realism and semantic consistency.

\section{Dataset Curation}
\label{sec:Dataset}
\begin{wrapfigure}{r}{0.45\columnwidth}
  \centering
  \vspace{-3mm}
  \includegraphics[width=0.45\columnwidth]{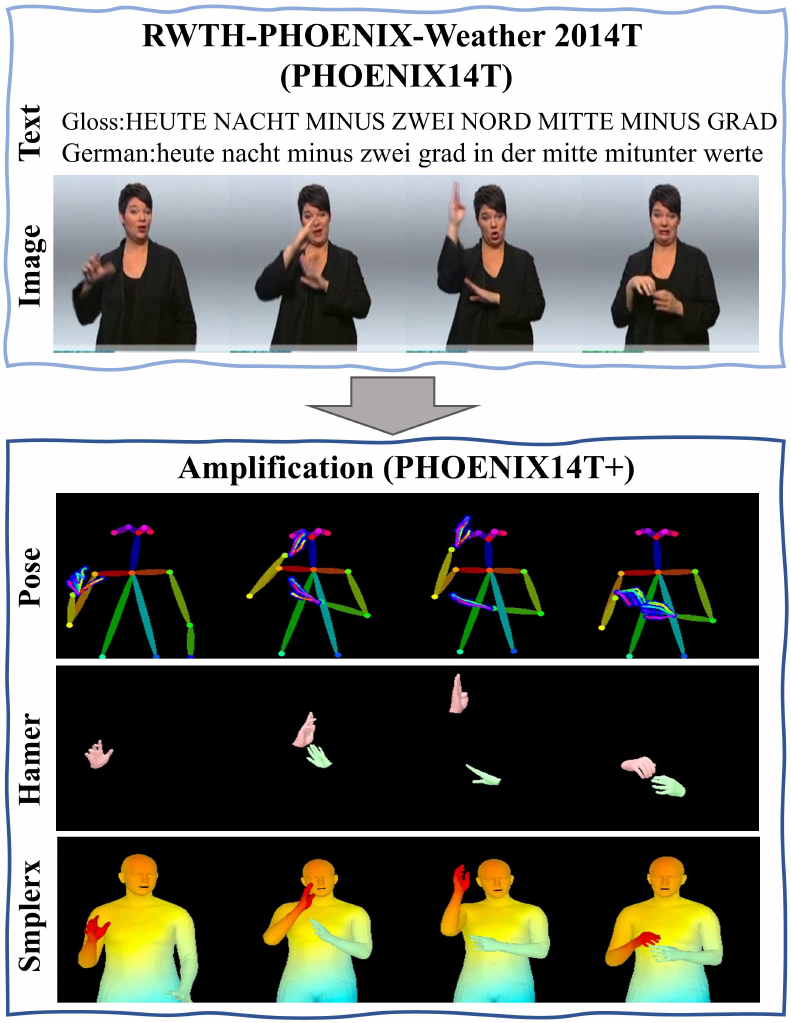}
  \caption{Some examples of the new dataset PHOENIX14T+.}
  \label{fig:dataset}
  \vspace{-2mm}
\end{wrapfigure}
Since the variety of datasets available for existing sign language generation tasks is relatively limited and monolithic in form, most of them are confined to only videos and the corresponding skeleton coordinates. Not only that, with the rapid development of large-scale, more and more demands pay more attention to the generation of real-life sign language videos, but only the skeleton form of the data can not do such an effect, lack of many details, can not meet the needs of reality. Therefore, we formally augment the widely used sign language dataset-German sign language corpus RWTH-PHOENIX-Weather 2014T (PHOENIX14T)~\cite{camgoz2018neural} with multiple pre-training models, including high-precision skeleton data (\textbf{Pose}) containing facial keypoints, 3D full-body pose (\textbf{Smplerx}), and (\textbf{Hamer}) reflecting hand details, aiming to provide the community with a novel and diverse large-scale sign language corpus suitable for both practical applications and academic research. In this section, the details of the dataset generation are elaborated. Figure ~\ref{fig:dataset} shows the three types of sign language representations and their corresponding raw video frames. \textbf{Pose} renders the human body gestures with color-differentiated key points and edges, \textbf{Hamer} renders the left and right hand joints with different colors to highlight the two-handed movements, and \textbf{Smplerx} renders the three-dimensional human body structures with gradient colors based on the depth information, which further enhances the three-dimensionality of the motion representation.

\textbf{Pose} We use an efficient whole-body state estimation model, DWPose~\cite{yang2023effective}, to extract the 2D coordinates of human joints, hands, and faces, totaling 60 keypoints, pose$\in R^{T\times 60\times 2}$. Unlike the traditional OpenPose~\cite{cao2017openpose} method, DWPose achieves lightweight deployment while maintaining high accuracy through an innovative two-phase distillation strategy combined with hierarchical knowledge transfer from a teacher-student model, which is widely used in computer vision, medical rehabilitation, and other fields. In particular, it shows higher accuracy in hand detail extraction, especially when dealing with large movements. 

\textbf{Hamer} In sign language video analysis, hand movements serve as the core semantic carriers, and their refined spatial representations are crucial for accurately conveying semantic information, especially the subtle deformations of the finger joints. Therefore, we adopt the state-of-the-art 3D gesture estimation method HaMeR~\cite{pavlakos2024reconstructing}, which is based on a transformer model trained on a large-scale hand dataset, and is able to provide high-quality 3D and depth information about the hand for more accurate understanding of the spatial structure of gestures. Specifically, we decode two sets of key parameters from the pre-trained model: the hand gesture parameter $\theta_\text{h} \in R^{16\times 3}$ (characterizing the joint rotations and displacements) and the shape parameter $\beta_\text{h} \in R^{10}$ (describing the morphological differences of individual hands). These parameters are mapped by the function $M(\theta_\text{h}, \beta_\text{h})$ to a high-resolution hand mesh $M \in R^{V\times 3}$, where $V=778$ vertices are connected by triangular facets to form a complete 3D hand model. Compared to Smplerx, Hamer has higher accuracy in complex gesture estimation.

\textbf{Smplerx} We estimate expressive human pose and shape parameters by means of a 3D parametric human model, SMPLer-X~\cite{cai2023smpler}. Specifically, we first estimate the pose parameters $\theta_\text{s} \in R^{55\times 3}$, including body, hand, eye, and chin poses; joint body, hand, and face shapes $\beta_\text{s} \in R^{10}$ as well as facial expressions $\psi \in R^{10}$ by pretraining the model, and then model the body, hand, and face geometries by combining the parameters through a joint regressor, which results in 10475 3D mesh vertices, and finally use rendering techniques to draw the Video. In this way, the rendered results simultaneously integrate 3D spatial, depth and continuous semantic information, thus effectively compensating for the lack of conventional skeletal data that contains only 2D information. In particular, this process eliminates the need to map the 2D skeleton to 3D space through an additional transformation step, avoiding possible errors.

\section{Method}
\label{sec:Method}
This work presents SignAligner, a novel method for generating real sign language videos from text. The method is divided into three stages: text-driven pose modalities co-generation, online collaborative correction of multimodality, and realistic sign video synthesis. First, a joint sign generator simultaneously produces posture coordinates, gesture actions, and body movements by incorporating text information. A Transformer-based text encoder extracts semantic features, while a cross-modal attention mechanism ensures accurate mapping and diversity control of sign language representations. Next, an online correction strategy, using cross-modal attention and dynamic loss weighting, refines the pose modalities, achieving information complementarity and resolving spatiotemporal conflicts for consistent and semantically coherent results. Finally, the corrected pose modalities are fed into a pre-trained video generation network to produce high-fidelity sign videos.

\subsection{Text-driven Pose Modalities Co-generation}
\subsubsection{Text Semantic Feature Extraction}

To extract the semantic features of gloss, we build a transformer-based encoder. A linear embedding layer is adopted to map glosses into a high-dimensional feature space. We further apply a positional coding layer to complement the temporal order of the gloss vectors. The computation is as follows:
\begin{eqnarray}\begin{aligned}
t'_\text{n}= W^t\cdot t_\text{n} + b^t + PE(\text{n}),\\
\end{aligned}\end{eqnarray}
where $t_\text{n}$ is a one-hot vector of the n-th gloss over the gloss vocabulary $\mathcal V$, $PE$ is conducted by the sine and cosine functions on the temporal text and pose order as in~\cite{vaswani2017transformer}. $W^t$ and $b^t$ represent the weight and bias respectively.
Next, we input the obtained gloss embeddings $\{t'\}_\text{{n=1}}^\text{N}$ into the gloss encoder to capture the global semantics of the glosses. Here, the encoder consists of n identical blocks, each including a Multi-Head Attention($MHA$), two Normalisation Layers ($NL$), and a Feedforward Layer ($FL$). The calculation process can be expressed as:
\begin{eqnarray}\begin{aligned}
\tilde{t}_\text{{1:N}}=GlossEncoder(t'_\text{{1:N}}).
\end{aligned}\label{eq:encoder}\end{eqnarray} 
The self-attention mechanism computes contextual dependencies using scaled dot-product attention:
\begin{equation}
\text{Attention}(Q, K, V) = \text{softmax}\left(\frac{QK^\top}{\sqrt{d}}\right)V,
\end{equation}
where $Q, K, V \in \mathbb{R}^{d \times d}$ are the query, key, and value matrices derived from $\mathbf{t}'_n$, and $d$ is the feature dimension.

\subsubsection{Pose Modalities Co-generation}
In this work, we aim to simultaneously generate pose modalities including \textbf{Pose}, \textbf{Hamer}, and \textbf{Smplerx} for temporal consistency. Therefore, similar to gloss encoding, we encode pose, hamer and smplerx as high-dimensional feature spaces through linear and positional encoding layers $PE_\text{p}$, $PE_\text{h}$ and $PE_\text{s}$, respectively. The computational formula is as follows:
\begin{eqnarray}\begin{aligned}
\left\{\begin{array}{l}
p'_\text{m} = W^p\cdot {p}_\text{m} + b^p + PE_\text{p}(\text{m});\\
h'_\text{m} = W^h\cdot {h}_\text{m} + b^h + PE_\text{h}(\text{m});\\
s'_\text{m} = W^s\cdot {s}_\text{m} + b^p + PE_\text{s}(\text{m}),\\
\end{array}\right.
\end{aligned}\end{eqnarray}
where $p_\text{m}$, $h_\text{m}$ and $s_\text{m}$ denote the coordinates of pose, hamer and smplerx at $m$-th timestamp, respectively. $W^p$, $W^h$, $W^s$ and $b^p$, $b^h$, $b^s$ denote the learnable weights and bias, respectively.

At the same time, in order to avoid spatio-temporal differences between different modalities, we construct a multi-model joint training framework for the PoseDecoder, HamerDecoder and SmplerxDecoder, respectively, and input visual features into the PoseDecoder, HamerDecoder and SmplerxDecoder constructed based on transformer, and construct a dynamic feedback and adjustment mechanism during the training process, so that we can generate the three kinds of sign language representations at one time. Take the pose decoder as an example, it can predict the next pose $\tilde{p}_\text{{m+1}}$ by aggregating all previously generated poses $\tilde{p}_\text{{1:m}}$. We also use an cross-attention mechanism in the transformer to enable semantic interaction between textual and visual sequences. The various decoders can be represented as:
\begin{eqnarray}\begin{aligned}
\tilde{p}_\text{{m+1}}=\text{PD}(p'_\text{{1:m}}, \tilde{t}_\text{{1:N}});\
\tilde{h}_\text{{m+1}}=\text{HD}(h'_\text{{1:m}}, \tilde{t}_\text{{1:N}});\
\tilde{s}_\text{{m+1}}=\text{SD}(s'_\text{{1:m}}, \tilde{t}_\text{{1:N}}),
\end{aligned}\end{eqnarray}
where PD, HD and SD stand for PoseDecoder, HamerDecoder and SmplerxDecoder respectively.

After $M$ time stamps, we have obtain the pose representation $\{\tilde{p}\}_\text{{m=1}}^\text{M}$, hamer representation $\{\tilde{h}\}_\text{{m=1}}^\text{M}$, smplerx representation $\{\tilde{s}\}_\text{{m=1}}^\text{M}$. In the training stage, the Mean Absolute Error (MAE) loss is used to constraint the consistency of the generated poses $\{\tilde{p}\}_\text{{m=1}}^\text{M}$, generated hamers $\{\tilde{h}\}_\text{{m=1}}^\text{M}$ and generated smplerxs $\{\tilde{s}\}_\text{{m=1}}^\text{M}$, respectively, with the ground truth $\widehat{P}=\{\widehat{p}\}_\text{{m=1}}^\text{M}$, $\widehat{H}=\{\widehat{h}\}_\text{{m=1}}^\text{M}$ and $\widehat{S}=\{\widehat{s}\}_\text{{m=1}}^\text{M}$.
\begin{eqnarray}
\mathcal L_\text{TMC} = \frac{1}{M}\sum_{m=1}^{M}(|\tilde{p}_\text{m}-\widehat{p}_\text{m}|+|\tilde{h}_\text{m}-\widehat{h}_\text{m}|+|\tilde{s}_\text{m}-\widehat{s}_\text{m}|).
\label{eq:accloss}
\end{eqnarray}

\subsection{Online Collaborative Correction}
Due to the heterogeneity of multimodal data, direct fusion may cause information inconsistency and noise, while using separate decoders for each modality can lead to incomplete learning of complex features. To address this, we propose an online multimodal correction strategy that dynamically adjusts the correlation between modalities, improving robustness and consistency in fusion and enabling more effective cross-modal interaction. 
We design a triple cross-modal attention pathway to focus spatial attention on hand features using the motion priors of skeleton data, enhancing the capture of motion details. Additionally, local hand motion semantics drive the spatio-temporal enhancement of the full-body video, improving the whole-body gesture’s spatio-temporal semantic representation. Dynamic complementarity between modalities is achieved by back-optimizing the joint confidence of skeleton sequences with global pose information, as expressed below: 
\begin{eqnarray}\begin{aligned}
\tilde{p'}_\text{1:m} = \text{CA}(\tilde{p}_\text{1:m},\tilde{h}_\text{1:m},\tilde{s}_\text{1:m});
\tilde{h'}_\text{1:m} = \text{CA}(\tilde{h}_\text{1:m},\tilde{p}_\text{1:m},\tilde{s}_\text{1:m});
\tilde{s'}_\text{1:m} = \text{CA}(\tilde{s}_\text{1:m},\tilde{p}_\text{1:m},\tilde{h}_\text{1:m}),\\
\end{aligned}\end{eqnarray}
where CA stand for cross-attention.

In the correction stage, to realize the effective fusion of multimodal cues, we design a dynamic loss weighting module to dynamically adjust the importance of each loss, learnable parameters $\alpha, \beta, \gamma \in \mathbb{R}^+$ are introduced and normalized by the softmax function in order to obtain the weight coefficients for each loss:
\begin{eqnarray}\begin{aligned}
w_A,w_B,w_C = softmax(\alpha, \beta, \gamma) = [\frac{e^\alpha}{e^\alpha + e^\beta + e^\gamma}, \frac{e^\beta}{e^\alpha + e^\beta + e^\gamma}, \frac{e^\gamma}{e^\alpha + e^\beta + e^\gamma}],
\end{aligned}\end{eqnarray}
where $w_{A}$, $w_{B}$ and $w_{C}$ are the weights of modes $Pose$, $Hamer$ and $Smplerx$ respectively.

During the training process, $\alpha, \beta, \gamma$ are learnable parameters, which are automatically updated by backpropagation to achieve dynamic synergy and adaptive optimization among different modalities.

Finally, to ensure the semantic consistency of multimodal features and the degree of complementarity among different modalities, we dynamically constrain the generative features of the three modalities through the obtained adaptive weights to ensure that the modified three sign language features are closer to the real semantics and promote semantic consistency among different modalities.
\begin{eqnarray}\begin{aligned}
\mathcal L_\text{OMC} = w_A \cdot ||\tilde{p'}_\text{m}-\widehat{p}_\text{m}||^2_{2} + w_B \cdot ||\tilde{h'}_\text{m}-\widehat{h}_\text{m}||^2_{2} +  w_C \cdot ||\tilde{s'}_\text{m}-\widehat{s}_\text{m}||^2_{2},
\end{aligned}\end{eqnarray}
where $\tilde{p'}_\text{{m=1}}^\text{M}$, $\tilde{h'}_\text{{m=1}}^\text{M}$ and $\tilde{s'}_\text{{m=1}}^\text{M}$ are generated features after calibration, and $\widehat{p}_\text{{m=1}}^\text{M}$, $\widehat{h}_\text{{m=1}}^\text{M}$ and $\widehat{s}_\text{{m=1}}^\text{M}$ are the corresponding real labels.

\subsection{Realistic Sign Video Synthesis}
In order to generate highly realistic sign language videos, we adopt the RealisDance~\cite{zhou2024realisdance} and retrain it with PHOENIX14T+ to meet the specific needs of sign language video generation.
First, we pass the corrected pose skeleton pose, hand video hamer, and full-body pose smplerx through a pose gating module, and integrate this information to fine-tune the RealiDance main framework. Since the SMPL-X model provides a unified and realistic human model by jointly training the shape parameters of the face, hand, and body, the resulting smplerx data ensures the coordination and realism of hand and body movements in sign language videos. The specific process can be expressed as follows:
\begin{eqnarray}\begin{aligned}
Video = RealisDance(\tilde{p'}_\text{{1:m}},\tilde{h'}_\text{{1:m}},\tilde{s'}_\text{{1:m}}).
\end{aligned}\end{eqnarray}

In the model training process, we define a joint loss function $L_\text{EVS}$, consisting of $L_\text{rec}$ and loss $L_\text{adv}$.
\begin{eqnarray}\begin{aligned}
L_\text{EVS} = L_\text{rec} + \lambda L_\text{adv},
\end{aligned}\end{eqnarray}
where the reconstruction loss $L_\text{rec}$ measures the difference between the generated video and the real video, the adversarial loss $L_\text{adv}$ improves the realism of the generated video by introducing a discriminator, and $\lambda$ is a weighting coefficient to balance the two. 


\begin{table*}[t]
\renewcommand\arraystretch{1.4}
\caption{Quantitative results on the PHOENIX14T+ dataset. 
}
\label{tab:Quantitative results on PHOENIX14T+ dataset}
\centering
\resizebox{\textwidth}{!}{
\begin{tabular}{cccccccc@{\hskip 0.1pt}cccccccc}
\Xhline{1pt}
   \multirow{2}{*}{Methods} &\multicolumn{6}{c}{DEV} &~ &\multicolumn{6}{c}{TEST}\\
   \cline{2-7}\cline{9-14}
    &BLEU-1{$\uparrow$} &BLEU-4{$\uparrow$} &ROUGE{$\uparrow$} &SSIM{$\uparrow$} &PSNR{$\uparrow$} &FID{$\downarrow$} &~ &BLEU-1{$\uparrow$} &BLEU-4{$\uparrow$} &ROUGE{$\uparrow$} &SSIM{$\uparrow$} &PSNR{$\uparrow$} &FID{$\downarrow$}\\
\Xhline{0.5pt}
    PTSLP + RealisDance$^\dag$~\cite{saunders2020ptslp,zhou2024realisdance}  &8.55 &1.68 &9.15 &0.58 &11.28 &51.64 &~ &8.86 &1.52 &8.83 &0.58 &11.45 &52.12 \\
    CogvideoX~\cite{yang2024cogvideox}  &8.14 &0.46 &7.21 &0.29 &3.82 &263.82 &~ &8.40 &0.51 &7.33 &0.29 &3.82 &264.75 \\
\Xhline{0.5pt}
    SignAligner (Ours) &{\bfseries 19.33} &{\bfseries 7.36} &{\bfseries 21.08} &{\bfseries 0.73} &{\bfseries 15.29} &{\bfseries 25.98} &~ &{\bfseries 20.56} &{\bfseries 8.17} &{\bfseries 20.88} &{\bfseries 0.73} &{\bfseries 15.32} &{\bfseries 26.26} \\
\Xhline{1pt}
\end{tabular}}
\end{table*}


\begin{figure}[t]
  \centering
  \includegraphics[width=\columnwidth]{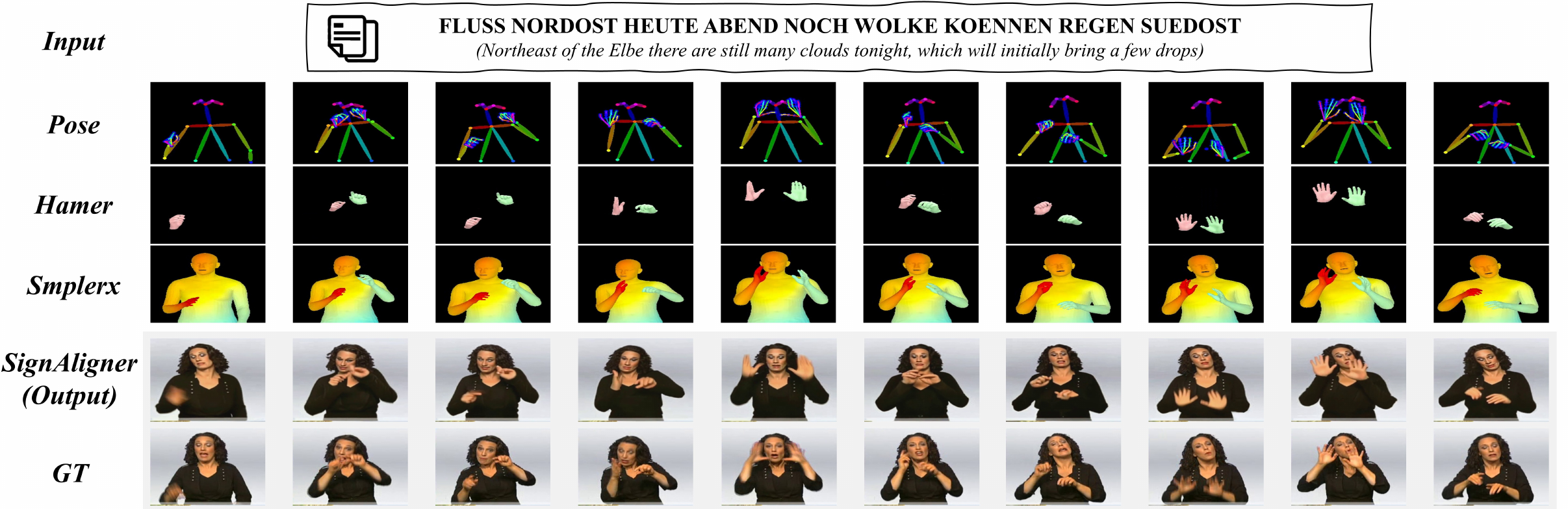}
  \caption{Visualization examples of produced sign language video sequence of SignAligner.}
  \label{fig:overline}
\end{figure}

\section{Experiments}
\label{sec:Experiments}

\subsection{Experimental Settings}
\textbf{Implementation Details.} 
In the text-driven pose modalities co-generation, we use a Transformer-based generative model with the Adam optimizer for training. For the OCC, the cross-modal multi-head attention mechanism is configured with 2 layers, 4 attention heads, and a batch size of 64. Finally, end-to-end video generation is achieved using the RealisDance~\cite{zhou2024realisdance}, retrained on the proposed PHOENIX14T+ dataset. All experiments were conducted on 8 NVIDIA GeForce RTX A6000 GPUs.

\noindent{\textbf{Evaluation Metrics.}} 
We use NSLT~\cite{camgoz2018neural} as an offline back-translation evaluation tool for pose, building on prior work~\cite{saunders2020ptslp, huang2021towards, tang2022gloss}. For Hamer, Smplerx, and video, we adopt the GFSLT method~\cite{zhou2023gloss} and train it on e PHOENIX14T+ to directly evaluate sign language videos. To assess the quality of our method, we use standard metrics in the sign language generation field: BLEU, ROUGE and \textit{Word Error Rate} (WER). For BLEU, we provide \textit{n}-grams from 1 to 4 for evaluating phase completeness. And to evaluate the quality of the generated video, we also use the \textit{Structural Similarity Index} (SSIM), \textit{Peak Signal-to-Noise Ratioand} (PSNR) and \textit{Fréchet Inception Distance} (FID) to measure the similarity between the generated and real videos.

\subsection{Comparison with State-of-the-Arts}
\begin{wrapfigure}{r}{0.45\columnwidth}
  \centering
  \vspace{-3mm}
  \includegraphics[width=0.45\columnwidth]{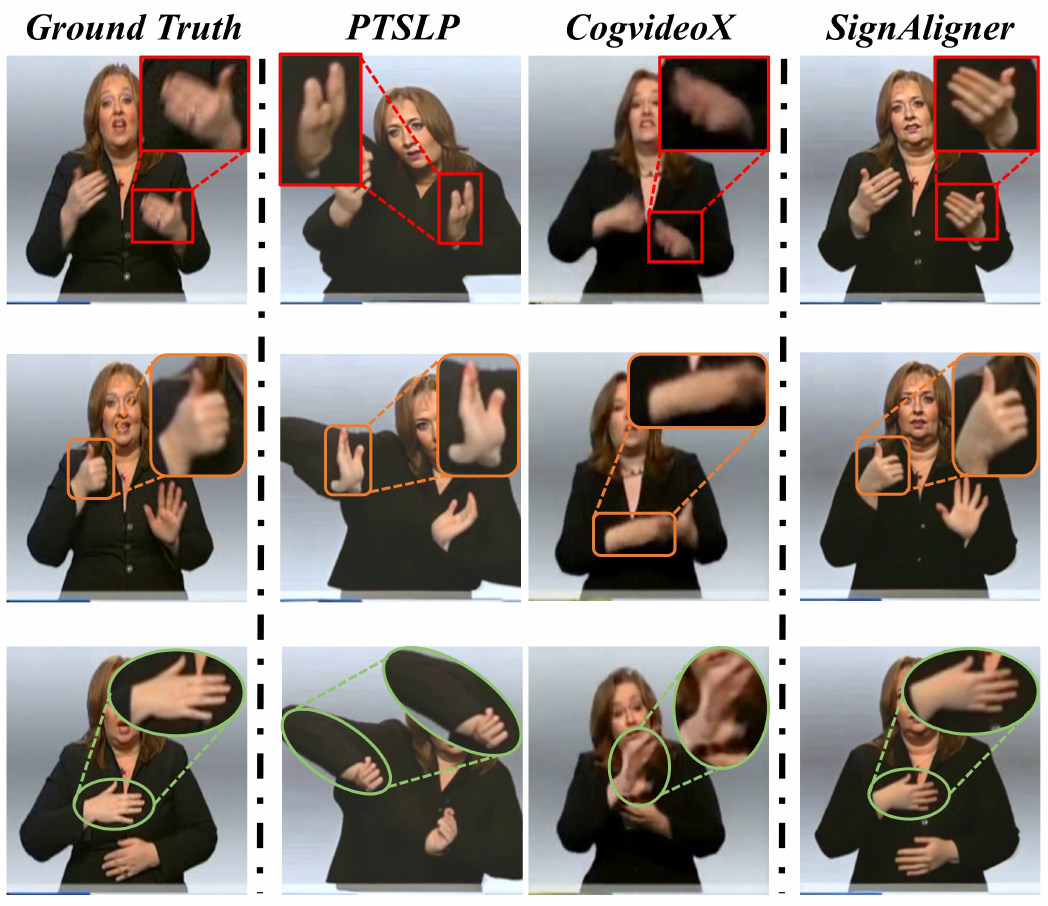}
  \caption{\small Visualization examples of PTSLP, CogvideoX and SignAligner on PHOENIX14T+.}
  \label{fig:result-1}
  \vspace{-3mm}
\end{wrapfigure}

As shown in Table~\ref{tab:Quantitative results on PHOENIX14T+ dataset}, the effectiveness of our multi-stage SLG method is demonstrated through comparisons with existing methods on the PHOENIX14T+ dataset, where it outperforms all other approaches. Compared to the sign language video generated by RealisDance\cite{zhou2024realisdance} based on the transformer baseline PTSLP\cite{saunders2020ptslp}, SignAligner shows significant improvements in all evaluation metrics, especially on the DEV/TEST sets for BLEU-1 and ROUGE, with performance gains of 10.78\% and 11.83\%, respectively. While the mainstream video generation model CogvideoX~\cite{yang2024cogvideox} achieves some results (8.40\% BLEU-1 on the test set), it struggles with the unique multimodal features of sign language, resulting in less accurate semantic mapping and suboptimal alignment between text and signing actions. 
Beyond semantic accuracy, SignAligner also significantly improves video quality. Compared to PTSLP + RealisDance, our method achieves higher SSIM and PSNR scores. Specifically, on the test set, SignAligner achieves SSIM of 0.73 and PSNR of 15.32, indicating better visual fidelity and clarity. These results highlight SignAligner's strengths in both language accuracy and visual quality, proving its robustness for high-fidelity sign language video generation. As shown in Figure~\ref{fig:overline}, the generated video sequence closely matches the ground truth.

We further provide a visualization example to show the performance differences of different methods in sign video generation, as shown in Figure~\ref{fig:result-1}. Compared to PTSLP and CogvideoX, SignAligner demonstrates significant advantages, particularly in hand structure, movement trajectory, and spatiotemporal coordination. PTSLP and CogvideoX often suffer from blurred hands, misaligned postures, and discontinuous movements, making gesture details and movement continuity difficult to restore. In contrast, SignAligner produces more natural hand movements and smoother transitions, closely matching the ground truth. This demonstrates SignAligner's effectiveness in improving clarity, accuracy, and consistency in generated sign language videos through multimodal collaborative modeling and realistic video synthesis.

\subsection{Ablation Study}

\noindent{\textbf{The impact of co-gen and OCC.}} To verify the effectiveness of the proposed pose modalities
\begin{wraptable}{r}{0.6\textwidth}
\renewcommand\arraystretch{1.4}
\vspace{-3mm}
\caption{\small Ablation results on PHOENIX14T+ dataset.}
\label{tab:Ablation results of different modules on the PHOENIX14T+.}
\centering
\resizebox{0.6\textwidth}{!}{
\begin{tabular}{cccccccc}
\Xhline{1pt}
    Methods &BLEU-1{$\uparrow$} &BLEU-4{$\uparrow$} &ROUGE{$\uparrow$} &SSIM{$\uparrow$} &PSNR{$\uparrow$} &FID{$\downarrow$}\\
\Xhline{0.5pt}
    w/o co-gen &14.50 &5.11 &15.54 &0.68 &13.79 &38.16 \\
    w/o OCC &17.84 &7.07 &19.01 &0.70 &14.67 &37.23 \\
\Xhline{0.5pt}
    Full (Ours) &{\bfseries 20.56} &{\bfseries 8.17} &{\bfseries 20.88} &{\bfseries 0.73} &{\bfseries 15.32} &{\bfseries 26.26} \\
\Xhline{1pt}
\vspace{-8mm}
\end{tabular}}
\end{wraptable}
co-generation \textbf{(co-gen)} and Online Collaborative Correction \textbf{(OCC)} mechanisms, we conduct ablation experiments on the PHOENIX14T+ dataset. As shown in Table~\ref{tab:Ablation results of different modules on the PHOENIX14T+.}, after removing Co-gen, BLEU-1 and BLEU-4 decreased to 14.50\% and 5.11\%, and SSIM and PSNR also significantly decreased, indicating that \textbf{(co-gen)} plays a key role in improving language accuracy and visual fidelity. Without this mechanism, the model finds it difficult to effectively integrate posture modal information, resulting in a significant decrease in the coherence and accuracy of generated videos. 
At the same time, removing \textbf{OCC} also led to a decline in performance: BLEU-1 and ROUGE decreased to 17.84\% and 19.01\%, and visual indicators also significantly decreased. 
This indicates that the real-time adjustment capability provided by \textbf{OCC} during the generation process is crucial for ensuring semantic consistency and visual quality. The complete model performed the best in all indicators, fully verifying the synergistic effect of \textbf{co-gen} and \textbf{OCC} in improving language expression and visual presentation.

\noindent{\textbf{Generate more accurate skeleton pose.}} 
Here, as shown in Table ~\ref{tab:Results on PHOENIX14T+ for Text to Pose task.}, our proposed SignAligner
\begin{wraptable}{r}{0.52\textwidth}
\renewcommand\arraystretch{1.2}
\vspace{-3mm}
\caption{\small Results on PHOENIX14T+ for Text to Pose task.}
\label{tab:Results on PHOENIX14T+ for Text to Pose task.}
\centering
\resizebox{0.52\textwidth}{!}{
\begin{tabular}{ccccc}
\Xhline{1pt}
    Methods &BLEU-1{$\uparrow$} &BLEU-4{$\uparrow$} &ROUGE{$\uparrow$} &WER{$\downarrow$} \\
\Xhline{0.5pt}
    PTSLP~\cite{saunders2020ptslp}  &13.35 &4.31 &13.17 &96.50 \\
    NAT-AT~\cite{huang2021towards} &14.26 &5.53 &18.72 &88.15 \\
    NAT-EA~\cite{huang2021towards} &15.12 &6.66 &19.43 &82.01 \\
    DET~\cite{viegas2023including} &17.18 &5.76 &17.64 &-- \\
    G2P-DDM~\cite{xie2024g2p} &16.11 &7.50 &-- &77.26 \\
    GCDM~\cite{tang2024GCDM} &22.03 &7.91 &23.20 &81.94 \\
    GEN-OBT~\cite{tang2022gloss} &23.08 &8.01 &23.49 &81.78 \\ 
\Xhline{0.5pt}
    Ours &{\bfseries 24.39} &{\bfseries 8.47} &{\bfseries 25.21} &{\bfseries 73.89} \\
\Xhline{1pt}
\vspace{-5mm}
\end{tabular}}
\end{wraptable}
achieves significantly better performance than existing models in the Text to Pose task, reaching BLEU-1 24.39\% and ROUGE 25.21\%. This shows that our method can not only effectively capture the semantic information in the text, but also generate action sequences with good language structure and coherence. The high semantic fidelity and sequence generation quality fully verify our design advantages in multimodal modeling and cross-modal alignment, and further prove the strong potential of SignAligner in text-driven sign language generation tasks.

\begin{table*}[h]
\renewcommand\arraystretch{1.4}
\vspace{-2mm}
\caption{\small Results on the PHOENIX14T+ dataset for Text to Hamer and Text to Smplerx tasks.}
\label{tab:Results on the PHOENIX14T+ dataset for Text to Hamer and Text to Smplerx tasks.}
\centering
\resizebox{\textwidth}{!}{
\begin{tabular}{ccccccc@{\hskip 0.1pt}ccccccc}
\Xhline{1pt}
   \multirow{2}{*}{Methods} &\multicolumn{5}{c}{Text to Hamer} &~ &\multicolumn{5}{c}{Text to Smplerx}\\
   \cline{2-6}\cline{8-12}
    &BLEU-1{$\uparrow$} &ROUGE{$\uparrow$} &SSIM{$\uparrow$} &PSNR{$\uparrow$} &FID{$\downarrow$} &~ &BLEU-1{$\uparrow$} &ROUGE{$\uparrow$} &SSIM{$\uparrow$} &PSNR{$\uparrow$} &FID{$\downarrow$}\\
\Xhline{0.5pt}
    PTSLP~\cite{saunders2020ptslp}  &13.26 &13.03 &0.95 &19.06 &25.65  &~ &9.89 &9.65 &0.79 &16.23 &7.58 \\
    CogvideoX~\cite{yang2024cogvideox}  &15.16 &14.17 &0.92 &14.93 &36.59 &~ &9.85 &7.67 &0.72 &12.03 &43.23 \\
\Xhline{0.5pt}
    Ours &{\bfseries 29.94} &{\bfseries 29.12} &{\bfseries 0.96} &{\bfseries 21.31} &{\bfseries 4.43} &~ &{\bfseries 27.48} &{\bfseries 27} &{\bfseries 0.83} &{\bfseries 18.65} &{\bfseries 3.61}\\
\Xhline{1pt}
\vspace{-5mm}
\end{tabular}}
\end{table*}

\noindent{\textbf{Capture fine finger details.}}
As shown in Table~\ref{tab:Results on the PHOENIX14T+ dataset for Text to Hamer and Text to Smplerx tasks.}, it can be seen that Ours achieved optimal performance in all indicators on Text to Hamer. Among them, BLEU-1 achieved 29.94\%, significantly better than the baseline method PTSLP (13.26\%), and compared to the popular text driven video model CogVideoX, we have achieved significant improvements. In addition, ours also demonstrated higher quality video generation performance in visual indicators, with SSIM and PSNR of 0.96 and 21.31, respectively, and FID reduced to 4.43, far superior to CogVideoX (FID of 36.59).

\noindent{\textbf{Maintain stronger body expressiveness.}}
Furthermore, we conducted the evaluation on Text to Smplerx. As shown in Table~\ref{tab:Results on the PHOENIX14T+ dataset for Text to Hamer and Text to Smplerx tasks.}, our model still achieved the best performance, especially significantly better than the traditional method PTSLP (9.89\%) on BLEU-1, significantly surpassing the performance of the current mainstream video generation model CogVideoX. At the same time, ours also performs outstandingly in perception indicators related to video quality. Achieved 0.83, 18.65 and 3.61 on the SSIM, PNSR and FID, respectively.

\section{Conclusions}
In this work, we introduce the PHOENIX14T+ dataset, an enhanced benchmark featuring three sign representations, Pose, Hamer, and Smplerx, to improve the expressiveness and realism of sign language generation. Building on this, we propose SignAligner, a novel three-stage method for high-quality sign language video generation. By leveraging a Transformer-based text encoder and a cross-modal attention mechanism, SignAligner effectively models semantic and motion relationships across pose modalities. The online collaborative correction of pose modalities further refines consistency and realism through dynamic loss adjustment. Finally, these refined representations are synthesized into accurate and realistic sign language videos. Experimental results show that SignAligner significantly outperforms existing methods in both accuracy and expressiveness.

\bibliographystyle{IEEEtran}
\bibliography{references}

\begin{thebibliography}{10}
\providecommand{\url}[1]{#1}
\csname url@samestyle\endcsname
\providecommand{\newblock}{\relax}
\providecommand{\bibinfo}[2]{#2}
\providecommand{\BIBentrySTDinterwordspacing}{\spaceskip=0pt\relax}
\providecommand{\BIBentryALTinterwordstretchfactor}{4}
\providecommand{\BIBentryALTinterwordspacing}{\spaceskip=\fontdimen2\font plus
\BIBentryALTinterwordstretchfactor\fontdimen3\font minus \fontdimen4\font\relax}
\providecommand{\BIBforeignlanguage}[2]{{%
\expandafter\ifx\csname l@#1\endcsname\relax
\typeout{** WARNING: IEEEtran.bst: No hyphenation pattern has been}%
\typeout{** loaded for the language `#1'. Using the pattern for}%
\typeout{** the default language instead.}%
\else
\language=\csname l@#1\endcsname
\fi
#2}}
\providecommand{\BIBdecl}{\relax}
\BIBdecl

\bibitem{Baldassarri2002SSL}
S.~Baldassarri, E.~Cerezo, and F.~Royo-Santas, ``Automatic translation system to spanish sign language with a virtual interpreter,'' vol. 5726, pp. 196--199, 2009.

\bibitem{glauert2006vanessa}
J.~Glauert, R.~Elliott, S.~Cox, J.~Tryggvason, and M.~Sheard, ``Vanessa – a system for communication between deaf and hearing people,'' \emph{Technology and Disability}, vol.~18, no.~4, pp. 207--216, 2006.

\bibitem{zelinka2019nn}
J.~Zelinka, J.~Kanis, and P.~Salajka, ``Nn-based czech sign language synthesis,'' in \emph{International Conference on Speech and Computer}, 2019, pp. 559--568.

\bibitem{cui2019deep}
R.~Cui, Z.~Cao, W.~Pan, C.~Zhang, and J.~Wang, ``Deep gesture video generation with learning on regions of interest,'' \emph{IEEE Transactions on Multimedia}, vol.~22, no.~10, pp. 2551--2563, 2019.

\bibitem{stoll2020text2sign}
S.~Stoll, N.~C. Camgoz, S.~Hadfield, and R.~Bowden, ``Text2sign: Towards sign language production using neural machine translation and generative adversarial networks,'' \emph{International Journal of Computer Vision}, vol. 128, no.~4, pp. 891--908, 2020.

\bibitem{krishna2021gan}
S.~Krishna and J.~Ukey, ``Gan based indian sign language synthesis,'' in \emph{Indian Conference on Vision, Graphics and Image Processing}, 2021, pp. 1--8.

\bibitem{hwang2021non}
E.~Hwang, J.-H. Kim, and J.-C. Park, ``Non-autoregressive sign language production with gaussian space,'' in \emph{British Machine Vision Conference}, 2021, pp. 1--13.

\bibitem{xiao2020skeleton}
Q.~Xiao, M.~Qin, and Y.~Yin, ``Skeleton-based chinese sign language recognition and generation for bidirectional communication between deaf and hearing people,'' \emph{Neural Networks}, vol. 125, pp. 41--55, 2020.

\bibitem{unanue2021berttune}
I.~J. Unanue, J.~Parnell, and M.~Piccardi, ``Berttune: Fine-tuning neural machine translation with bertscore,'' in \emph{Association for Computational Linguistics}, 2021, pp. 915--924.

\bibitem{radford2021clip}
A.~Radford, J.~W. Kim, C.~Hallacy, A.~Ramesh, G.~Goh, S.~Agarwal, G.~Sastry, A.~Askell, P.~Mishkin, J.~Clark \emph{et~al.}, ``Learning transferable visual models from natural language supervision,'' in \emph{International Conference on Machine Learning}, 2021, pp. 8748--8763.

\bibitem{saunders2020ptslp}
B.~Saunders, N.~C. Camgoz, and R.~Bowden, ``Progressive transformers for end-to-end sign language production,'' in \emph{European Conference on Computer Vision}, 2020, pp. 687--705.

\bibitem{huang2021towards}
W.~Huang, W.~Pan, Z.~Zhao, and Q.~Tian, ``Towards fast and high-quality sign language production,'' in \emph{ACM International Conference on Multimedia}, 2021, pp. 3172--3181.

\bibitem{wang2024linguistics}
X.~Wang, S.~Tang, P.~Song, S.~Wang, D.~Guo, and R.~Hong, ``Linguistics-vision monotonic consistent network for sign language production,'' in \emph{International Conference on Acoustics, Speech, and Signal Processing}, 2025.

\bibitem{tang2024signidd}
S.~Tang, J.~He, D.~Guo, Y.~Wei, F.~Li, and R.~Hong, ``Sign-idd: Iconicity disentangled diffusion for sign language production,'' in \emph{AAAI Conference on Artificial Intelligence}, vol.~39, no.~7, 2025, pp. 7266--7274.

\bibitem{fu2025speculative}
J.~Fu, Y.~Jiang, J.~Chen, J.~Fan, X.~Geng, and X.~Yang, ``Speculative ensemble: Fast large language model ensemble via speculation,'' \emph{arXiv preprint arXiv:2502.01662}, 2025.

\bibitem{li2022pre}
S.~Li, X.~Puig, C.~Paxton, Y.~Du, C.~Wang, L.~Fan, T.~Chen, D.-A. Huang, E.~Aky{\"u}rek, A.~Anandkumar \emph{et~al.}, ``Pre-trained language models for interactive decision-making,'' \emph{Neural Information Processing Systems}, vol.~35, pp. 31\,199--31\,212, 2022.

\bibitem{hu2023survey}
L.~Hu, Z.~Liu, Z.~Zhao, L.~Hou, L.~Nie, and J.~Li, ``A survey of knowledge enhanced pre-trained language models,'' \emph{IEEE Transactions on Knowledge and Data Engineering}, vol.~36, no.~4, pp. 1413--1430, 2023.

\bibitem{saunders2022signing}
B.~Saunders, N.~C. Camgoz, and R.~Bowden, ``Signing at scale: Learning to co-articulate signs for large-scale photo-realistic sign language production,'' in \emph{Proceedings of the IEEE/CVF Conference on Computer Vision and Pattern Recognition}, 2022, pp. 5141--5151.

\bibitem{xie2024sign}
P.~Xie, T.~Peng, Y.~Du, and Q.~Zhang, ``Sign language production with latent motion transformer,'' in \emph{Proceedings of the IEEE/CVF Winter Conference on Applications of Computer Vision}, 2024, pp. 3024--3034.

\bibitem{camgoz2018neural}
N.~C. Camgoz, S.~Hadfield, O.~Koller, H.~Ney, and R.~Bowden, ``Neural sign language translation,'' in \emph{Computer Vision and Pattern Recognition}, 2018, pp. 7784--7793.

\bibitem{tang2024GCDM}
S.~Tang, F.~Xue, J.~Wu, S.~Wang, and R.~Hong, ``Gloss-driven conditional diffusion models for sign language production,'' \emph{ACM Transactions on Multimedia Computing, Communications and Applications}, vol.~21, no.~4, pp. 1--17, 2025.

\bibitem{zelinka2020neural}
J.~Zelinka and J.~Kanis, ``Neural sign language synthesis: Words are our glosses,'' in \emph{Winter Conference on Applications of Computer Vision}, 2020, pp. 3395--3403.

\bibitem{vasani2020generation}
N.~Vasani, P.~Autee, S.~Kalyani, and R.~Karani, ``Generation of indian sign language by sentence processing and generative adversarial networks,'' in \emph{International Conference on Information Systems Security}, 2020, pp. 1250--1255.

\bibitem{peng2025lmm}
Y.~Peng, G.~Zhang, M.~Zhang, Z.~You, J.~Liu, Q.~Zhu, K.~Yang, X.~Xu, X.~Geng, and X.~Yang, ``Lmm-r1: Empowering 3b lmms with strong reasoning abilities through two-stage rule-based rl,'' \emph{arXiv preprint arXiv:2503.07536}, 2025.

\bibitem{yang2023exploring}
X.~Yang, Y.~Wu, M.~Yang, H.~Chen, and X.~Geng, ``Exploring diverse in-context configurations for image captioning,'' \emph{Neural Information Processing Systems}, vol.~36, pp. 40\,924--40\,943, 2023.

\bibitem{xie2024g2p}
P.~Xie, Q.~Zhang, P.~Taiying, H.~Tang, Y.~Du, and Z.~Li, ``G2p-ddm: Generating sign pose sequence from gloss sequence with discrete diffusion model,'' in \emph{Association for the Advancement of Artificial Intelligence}, 2024, pp. 6234--6242.

\bibitem{yin2024t2s}
A.~Yin, H.~Li, K.~Shen, S.~Tang, and Y.~Zhuang, ``T2s-gpt: Dynamic vector quantization for autoregressive sign language production from text,'' \emph{arXiv preprint arXiv:2406.07119}, 2024.

\bibitem{baltatzis2024neural}
V.~Baltatzis, R.~A. Potamias, E.~Ververas, G.~Sun, J.~Deng, and S.~Zafeiriou, ``Neural sign actors: a diffusion model for 3d sign language production from text,'' in \emph{Proceedings of the IEEE/CVF Conference on Computer Vision and Pattern Recognition}, 2024, pp. 1985--1995.

\bibitem{kaur2016hamnosys}
K.~Kaur and P.~Kumar, ``Hamnosys to sigml conversion system for sign language automation,'' \emph{Procedia Computer Science}, vol.~89, pp. 794--803, 2016.

\bibitem{brock2020learning}
H.~Brock, F.~Law, K.~Nakadai, and Y.~Nagashima, ``Learning three-dimensional skeleton data from sign language video,'' \emph{ACM Transactions on Intelligent Systems and Technology (TIST)}, vol.~11, no.~3, pp. 1--24, 2020.

\bibitem{stoll2018sign}
S.~Stoll, N.~C. Camg{\"o}z, S.~Hadfield, and R.~Bowden, ``Sign language production using neural machine translation and generative adversarial networks,'' in \emph{Proceedings of the 29th British Machine Vision Conference (BMVC 2018)}.\hskip 1em plus 0.5em minus 0.4em\relax British Machine Vision Association, 2018.

\bibitem{yang2023effective}
Z.~Yang, A.~Zeng, C.~Yuan, and Y.~Li, ``Effective whole-body pose estimation with two-stages distillation,'' in \emph{Proceedings of the IEEE/CVF International Conference on Computer Vision}, 2023, pp. 4210--4220.

\bibitem{cao2017openpose}
Z.~Cao, T.~Simon, S.-E. Wei, and Y.~Sheikh, ``Realtime multi-person 2d pose estimation using part affinity fields,'' in \emph{Computer Vision and Pattern Recognition}, 2017, pp. 7291--7299.

\bibitem{pavlakos2024reconstructing}
G.~Pavlakos, D.~Shan, I.~Radosavovic, A.~Kanazawa, D.~Fouhey, and J.~Malik, ``Reconstructing hands in 3d with transformers,'' in \emph{Proceedings of the IEEE/CVF Conference on Computer Vision and Pattern Recognition}, 2024, pp. 9826--9836.

\bibitem{cai2023smpler}
Z.~Cai, W.~Yin, A.~Zeng, C.~Wei, Q.~Sun, W.~Yanjun, H.~E. Pang, H.~Mei, M.~Zhang, L.~Zhang \emph{et~al.}, ``Smpler-x: Scaling up expressive human pose and shape estimation,'' \emph{Neural Information Processing Systems}, vol.~36, pp. 11\,454--11\,468, 2023.

\bibitem{vaswani2017transformer}
A.~Vaswani, N.~Shazeer, N.~Parmar, J.~Uszkoreit, L.~Jones, A.~N. Gomez, {\L}.~Kaiser, and I.~Polosukhin, ``Attention is all you need,'' in \emph{Neural Information Processing Systems}, 2017, pp. 1--11.

\bibitem{zhou2024realisdance}
J.~Zhou, B.~Wang, W.~Chen, J.~Bai, D.~Li, A.~Zhang, H.~Xu, M.~Yang, and F.~Wang, ``Realisdance: Equip controllable character animation with realistic hands,'' \emph{arXiv preprint arXiv:2409.06202}, 2024.

\bibitem{yang2024cogvideox}
Z.~Yang, J.~Teng, W.~Zheng, M.~Ding, S.~Huang, J.~Xu, Y.~Yang, W.~Hong, X.~Zhang, G.~Feng \emph{et~al.}, ``Cogvideox: Text-to-video diffusion models with an expert transformer,'' \emph{arXiv preprint arXiv:2408.06072}, 2024.

\bibitem{tang2022gloss}
S.~Tang, R.~Hong, D.~Guo, and M.~Wang, ``Gloss semantic-enhanced network with online back-translation for sign language production,'' in \emph{ACM International Conference on Multimedia}, 2022, pp. 5630--5638.

\bibitem{zhou2023gloss}
B.~Zhou, Z.~Chen, A.~Clap{\'e}s, J.~Wan, Y.~Liang, S.~Escalera, Z.~Lei, and D.~Zhang, ``Gloss-free sign language translation: Improving from visual-language pretraining,'' in \emph{Proceedings of the IEEE/CVF International Conference on Computer Vision}, 2023, pp. 20\,871--20\,881.

\bibitem{viegas2023including}
C.~Viegas, M.~Inan, L.~Quandt, and M.~Alikhani, ``Including facial expressions in contextual embeddings for sign language generation,'' in \emph{Joint Conference on Lexical and Computational Semantics}, 2023, pp. 1--10.

\end{thebibliography}

\end{document}